\RequirePackage[hyphens]{url}
\documentclass{article}
\usepackage{authblk}
\usepackage[margin=1.4in]{geometry}

\newcommand\fref{Figure~\ref}
\newcommand\tref{Table~\ref}

\usepackage{xspace}
\usepackage{graphicx}
\usepackage{multirow}
\usepackage{hhline}
\usepackage{comment}
\usepackage{adjustbox}
\usepackage{xcolor}
\usepackage{pifont}
\usepackage{subfigure}
\usepackage{tabulary}
\usepackage{array}
\usepackage{colortbl}
\usepackage{color,soul}
\usepackage[colorlinks=true,
citecolor=blue,
filecolor=black,
linkcolor=blue,
urlcolor=blue]{hyperref}
\usepackage{booktabs} 

\renewcommand{\baselinestretch}{1.0}
\usepackage[switch,columnwise]{lineno} 

\usepackage{amsmath}
\usepackage{amsfonts}       
\usepackage{dsfont}

\newcommand\extrafootertext[1]{%
    \bgroup
    \renewcommand\thefootnote{\fnsymbol{footnote}}%
    \renewcommand\thempfootnote{\fnsymbol{mpfootnote}}%
    \footnotetext[0]{#1}%
    \egroup
}
\usepackage{listings}

\title{DeepSpeed-Chat: Easy, Fast and Affordable RLHF Training of ChatGPT-like Models at All Scales}
\author[]{Zhewei Yao, Reza Yazdani Aminabadi, Olatunji Ruwase, Samyam Rajbhandari, Xiaoxia Wu, Ammar Ahmad Awan, Jeff Rasley, Minjia Zhang, Conglong Li, Connor Holmes, Zhongzhu Zhou, Michael Wyatt, Molly Smith, Lev Kurilenko, Heyang Qin, Masahiro Tanaka, Shuai Che, Shuaiwen Leon Song, Yuxiong He}
\affil[]{Deepspeed of Microsoft}

\date{}

\begin{document}

\maketitle

\begin{abstract}
ChatGPT-like models have revolutionized various applications in artificial intelligence, from summarization and coding to translation, matching or even surpassing human performance. However, the current landscape lacks an accessible, efficient, and cost-effective end-to-end RLHF (Reinforcement Learning with Human Feedback) training pipeline for these powerful models, particularly when training at the scale of billions of parameters. This paper introduces DeepSpeed-Chat, a novel system that democratizes RLHF training, making it accessible to the AI community. DeepSpeed-Chat offers three key capabilities: an easy-to-use training and inference experience for ChatGPT-like models, a DeepSpeed-RLHF pipeline that replicates the training pipeline from InstructGPT, and a robust DeepSpeed-RLHF system that combines various optimizations for training and inference in a unified way. The system delivers unparalleled efficiency and scalability, enabling training of models with hundreds of billions of parameters in record time and at a fraction of the cost. With this development, DeepSpeed-Chat paves the way for broader access to advanced RLHF training, even for data scientists with limited resources, thereby fostering innovation and further development in the field of AI.
\end{abstract}

\section{Overview}
ChatGPT~\cite{chatgpt} like models have taken the AI world by storm, and it would not be an overstatement to say that it's impact on the digital world has been revolutionary. 
These models are incredibly versatile, capable of performing tasks like summarization, coding, and translation with results that are on-par or even exceeding the capabilities of human experts. 
Given the sheer power of these models, multiple efforts are underway in the AI open-source community to make ChatGPT-style models more accessible (e.g. ChatLLaMa~\cite{chatllama}, Alpaca~\cite{alpaca}, Vicuna~\cite{zheng2023judging}, Databricks-Dolly~\cite{dolly}, etc.).

Despite these incredible efforts, there is still a lack of an end-to-end RLHF pipeline capable of training powerful ChatGPT like model that is easily accessible to the AI community. 
For instance, training a modest 6.7B ChatGPT model with existing systems typically requires expensive multi-GPU setup that is beyond the reach of many data scientists.
Even with access to such computing resources, training efficiency is often less than 5\% of what these machines are capable of (as illustrated in~\fref{fig:figure-6}). 
And finally, existing solutions simply cannot support easy, fast and affordable training state-of-the-art ChatGPT models with hundreds of billions of parameters, even given access to multi-GPU clusters.

These limitations stem from a lack of a robust system design that is capable of effectively supporting the complex InstructGPT’s RLHF training pipeline that is quite different from the standard pre-training and fine-tuning pipelines that existing DL systems are designed for. 
Therefore, in the spirit of democratizing ChatGPT-like models, and making RLHF training truly accessible to the AI community, today we are releasing DeepSpeed-Chat with the following three capabilities:
\begin{itemize}
\item  \textbf{Easy-to-use Training and Inference Experience for ChatGPT Like Models}: A single script capable of taking a pre-trained Huggingface~\cite{wolf2019huggingface} model, running it through all three steps of InstructGPT~\cite{ouyang2022training} training using DeepSpeed-RLHF system and producing your very own ChatGPT like model. 
In addition, we provide an inference API for testing conversation-style interactions after the model is trained.

\item \textbf{DeepSpeed-RLHF Pipeline}: DeepSpeed-RLHF pipeline primarily replicates the training pipeline from the InstructGPT~\cite{ouyang2022training} paper with careful attention to ensure completeness and one-to-one correspondence with the three-steps that includes a) Supervised Fine-tuning (SFT), b) Reward Model Fine-tuning and c) Reinforcement Learning with Human Feedback (RLHF)~\cite{stiennon2020learning}. 
Additionally, we offer data abstraction and blending capabilities to enable training with multiple data sources.

\item {DeepSpeed-RLHF System}: A robust and sophisticated RLHF system that combines the training and inference prowess of DeepSpeed into single unified Hybrid Engine (DeepSpeed-HE) for RLHF. 
The Hybrid-Engine is capable of seamlessly transitioning between inference and training modes within RLHF, allowing it to leverage various optimizations from DeepSpeed-Inference such as tensor-parallelism and high-performance transformer kernels for generation, while also benefiting from the multitude of ZeRO- and LoRA~\cite{hu2021lora}-based memory optimization strategies for RL training. 
DeepSpeed-HE is also aware of the full RLHF pipeline, allowing it to make optimal decisions in terms of memory management and data movement across different phases of RLHF.
\end{itemize}
DeepSpeed-RLHF system is capable of unparalleled efficiency at scale, making complex RLHF training fast, affordable, and easily accessible to the AI community:

\textbf{Efficiency and Affordability}: In terms of efficiency, DeepSpeed-HE is over 15x faster than existing systems, making RLHF training both fast and affordable. 
For instance, DeepSpeed-HE can train an OPT-13B~\cite{zhang2022opt} in just 9 hours and OPT-30B in 18 hours on Azure Cloud for under \$300 and \$600, respectively, as shown in~\tref{tab:table-1}.

\begin{table}[t]
\caption{
Single-Node 8x A100: Training Time and Corresponding Approximate Cost on Azure.
}
\label{tab:table-1}
\begin{adjustbox}{width=0.999\linewidth}
\centering
\begin{tabular}{lcccccc }
\toprule
GPUs   & OPT-6.7B &OPT-13B     & OPT-30B     & OPT-66B  \\
\midrule
8x A100-40GB    & 5.7 hours & 10.8 hours &	 1.85 days &	 NA \\
8x A100-80GB    & 4.1 hours (\$132) & 9 hours (\$290) & 18 hours (\$580) & 	 2.1 days (\$1620) \\
\bottomrule
\end{tabular}
\end{adjustbox}
\end{table}

\textbf{Excellent Scalability}: DeepSpeed-HE supports models with hundreds of billions of parameters and can achieve excellent scalability on multi-node multi-GPU systems. 
As a result, even a 13B model can be trained in 1.25 hours and a massive 175B model can be trained with DeepSpeed-HE in under a day as shown in~\tref{tab:table-2}.\footnote{
\textbf{Very Important Details}: The numbers in both tables (\ref{tab:table-1}, \ref{tab:table-2}) above are for Step 3 of the training and based on actual measured training throughput on DeepSpeed-RLHF curated dataset and training recipe which trains for one epoch on a total of 135M tokens. 
We have in total 67.5M query tokens (131.9k queries with sequence length 256) and 67.5M generated tokens (131.9k answers with sequence length 256), and a maximum global batch size per step of 0.5M tokens (1024 query-answer pairs). 
We urge readers to pay attention to these specifications before making any cost and e2e time comparisons with DeepSpeed-RLHF. 
See our benchmark settings (\url{https://github.com/microsoft/DeepSpeedExamples/tree/master/applications/DeepSpeed-Chat/training/step3_rlhf_finetuning/BenckmarkSetting.md}) page for more details.
}

\begin{table}[t]
\caption{
Multi-Node 64x A100-80GB: Training Time and Corresponding Approximate Cost on Azure.
}
\label{tab:table-2}
\begin{adjustbox}{width=0.999\linewidth}
\centering
\begin{tabular}{lcccccc }
\toprule
GPUs   &OPT-13B     & OPT-30B     & OPT-66B & OPT-175B  \\
\midrule
64x A100-80G	& 1.25 hours (\$320)	& 4 hours (\$1024) & 7.5 hours (\$1920)	& 20 hours (\$5120)\\
\bottomrule
\end{tabular}
\end{adjustbox}
\end{table}

\textbf{Democratizing RLHF Training}: With just a single GPU, DeepSpeed-HE supports training models with over 13 billion parameters as shown in~\tref{tab:table-3}, enabling data scientists without access to multi-GPU systems to create not just toy RLHF models but large and powerful ones that can be used in real-world scenarios. 

\begin{table}[t]
\caption{
Max Model Size Supported by DeepSpeed-HE on a Single GPU.
}
\label{tab:table-3}
\begin{adjustbox}{width=0.999\linewidth}
\centering
\begin{tabular}{lcccccc }
\toprule
     &	V100 32G & 	A6000 48G & A100 40G & A100 80G \\
\midrule
 Model Size&	OPT-2.7B & 	OPT-6.7B & OPT-6.7B & OPT-13B \\
\bottomrule
\end{tabular}
\end{adjustbox}
\end{table}

Next, we dive deeper into the three capabilities of DeepSpeed-Chat introduced above.

\label{sec:overview}

\section{Easy-to-use ChatGPT Training and Inference Experience}
\label{sec:easy-to-use}
We start with the easy-to-use experience by showing how you can train OPT-13B and then OPT-66B models with DeepSpeed-RLHF system. 
If you are short on time, you can even train an OPT-1.3B model on a single consumer-grade GPU in just two hours.  
We also demonstrate how you can use our DeepSpeed-chat RLHF API to develop your own custom pipelines.

\subsection{Training your first ChatGPT-Style model is so easy with DeepSpeed-Chat’s RLHF examples}
\paragraph{One single script completes all three stages of RLHF training and generates your first ChatGPT model!}
We use an example of pretrained OPT-13B as the actor model and OPT-350M as the reward model in the following single script to generate a final 13B ChatGPT-style model:

\begin{lstlisting}
pip install deepspeed>=0.9.0

git clone https://github.com/microsoft/DeepSpeedExamples.git
cd DeepSpeedExamples/applications/DeepSpeed-Chat/
pip install -r requirements.txt

python train.py --actor-model facebook/opt-13b --reward-model \
facebook/opt-350m --deployment-type single_node
\end{lstlisting}

In about half a day, your 13B model would be fully trained with its checkpoints ready.  
Table~\ref{tab:table-4} demonstrates a breakdown of the training time for each of the three steps.

\begin{table}[t]
\caption{
E2E time breakdown for training a 13 billion parameter ChatGPT model via DeepSpeed-Chat on a single DGX node with 8 NVIDIA A100-40G GPUs.
}
\label{tab:table-4}
\begin{adjustbox}{width=0.999\linewidth}
\centering
\begin{tabular}{lcccccc }
\toprule
Model Sizes                    & Step 1 & Step 2 & Step 3 & Total \\
\midrule
Actor: OPT-13B, Reward: OPT-350M& 2.5hr	& 0.25hr & 10.8hr & 13.6hr \\
\bottomrule
\end{tabular}
\end{adjustbox}
\end{table}

\paragraph{Test your final model with DeepSpeed-Chat’s Inference API}
Now it is time to plug-in and test your final model trained by DeepSpeed-Chat! You can play with not only Q\&A sessions but also conversation-style interactions.
\begin{lstlisting}
Human:     Do you know Microsoft?
Assistant: Yes, I'm a big fan of Microsoft. What about them?
Human:     Can you explain it to a 6-year old child? I wonder how 
           I should describe it
Assistant: Sure, I can try. Microsoft is a company that makes 
           computers, and they make a program called ``Windows'' 
           which is the operating system that runs on the computer. 
           It's like the ``brain'' of the computer. It's where 
           all the programs and files are stored. It's also where 
           you'll find the ``start'' button, which is where 
           you'll start to use the computer.
\end{lstlisting}

\subsection{Want to try different model sizes and configurations? You got it!}
We understand users often like to play with different model sizes and configurations to meet their training time, resources, and quality requirements. 
With DeepSpeed-Chat, users can easily do that. For example, if you want to train a larger and higher-quality model on your GPU cluster for your research or business, you can simply use the same script with your desired model size e.g., 66B and GPU counts e.g., 64 GPUs:
\begin{lstlisting}
python train.py --actor-model facebook/opt-66b \
--reward-model facebook/opt-350m --deployment-type multi_node
\end{lstlisting}
Within 9 hours, you can have your 66 billion parameters ChatGPT model ready to be served in your favorite front-end GUI as shown in~\tref{tab:table-5}.

\begin{table}[t]
\caption{
E2E time breakdown for training a 66 billion parameter ChatGPT model via DeepSpeed-Chat on 8 DGX nodes with 8 NVIDIA A100-80G GPUs/node.
}
\label{tab:table-5}
\begin{adjustbox}{width=0.999\linewidth}
\centering
\begin{tabular}{lcccccc }
\toprule
Model Sizes                    & Step 1 & Step 2 & Step 3 & Total \\
\midrule
Actor: OPT-66B, Reward: OPT-350M & 82 mins & 5 mins & 7.5hr  & 9hr \\
\bottomrule
\end{tabular}
\end{adjustbox}
\end{table}

If you only have around 1-2 hours for coffee or lunch break, you can also try to train a small/toy model with DeepSpeed-Chat. 
For example, we prepared a training example for a 1.3B model with a single dataset to test our framework on your consumer-grade GPUs. 
The best part is that you will have your model checkpoint ready to play with when you are back from your lunch break (see~\tref{tab:table-6})!
\begin{lstlisting}
python train.py --actor-model facebook/opt-1.3b \
--reward-model facebook/opt-350m --deployment-type single_gpu
\end{lstlisting}

\begin{table}[t]
\caption{
E2E time breakdown for training a 1.3 billion parameter ChatGPT model via DeepSpeed-Chat on a single commodity NVIDIA A6000 GPU with 48GB memory.}
\label{tab:table-6}
\begin{adjustbox}{width=0.999\linewidth}
\centering
\begin{tabular}{lcccccc }
\toprule
Model Sizes                    & Step 1 & Step 2 & Step 3 & Total \\
\midrule
 Actor: OPT-1.3B, Reward: OPT-350M & 2900 secs & 670 secs & 1.2hr & 2.2hr \\
\bottomrule
\end{tabular}
\end{adjustbox}
\end{table}

\subsection{Customizing your own RLHF training pipeline using DeepSpeed-Chat’s RLHF APIs}
DeepSpeed-Chat allows users to build their very own RLHF training pipeline using our flexible APIs shown below, which users can use to reconstruct their own RLHF training strategy. This enables a general interface and backend for creating a wide range of RLHF algorithms for research exploration.

\begin{lstlisting}
engine = DeepSpeedRLHFEngine(
  actor_model_name_or_path=args.actor_model_name_or_path,
  critic_model_name_or_path=args.critic_model_name_or_path,
  tokenizer=tokenizer,
  num_total_iters=num_total_iters,
  args=args)

trainer = DeepSpeedPPOTrainer(engine=engine, args=args)

for prompt_batch in prompt_train_dataloader:
  out = trainer.generate_experience(prompt_batch)
  actor_loss, critic_loss = trainer.train_rlhf(out)
\end{lstlisting}

\section{Full-fledged RLHF Training Pipeline}
\label{sec:full-rlhf}
\begin{figure}
\centering
\includegraphics[width=1.0\linewidth]{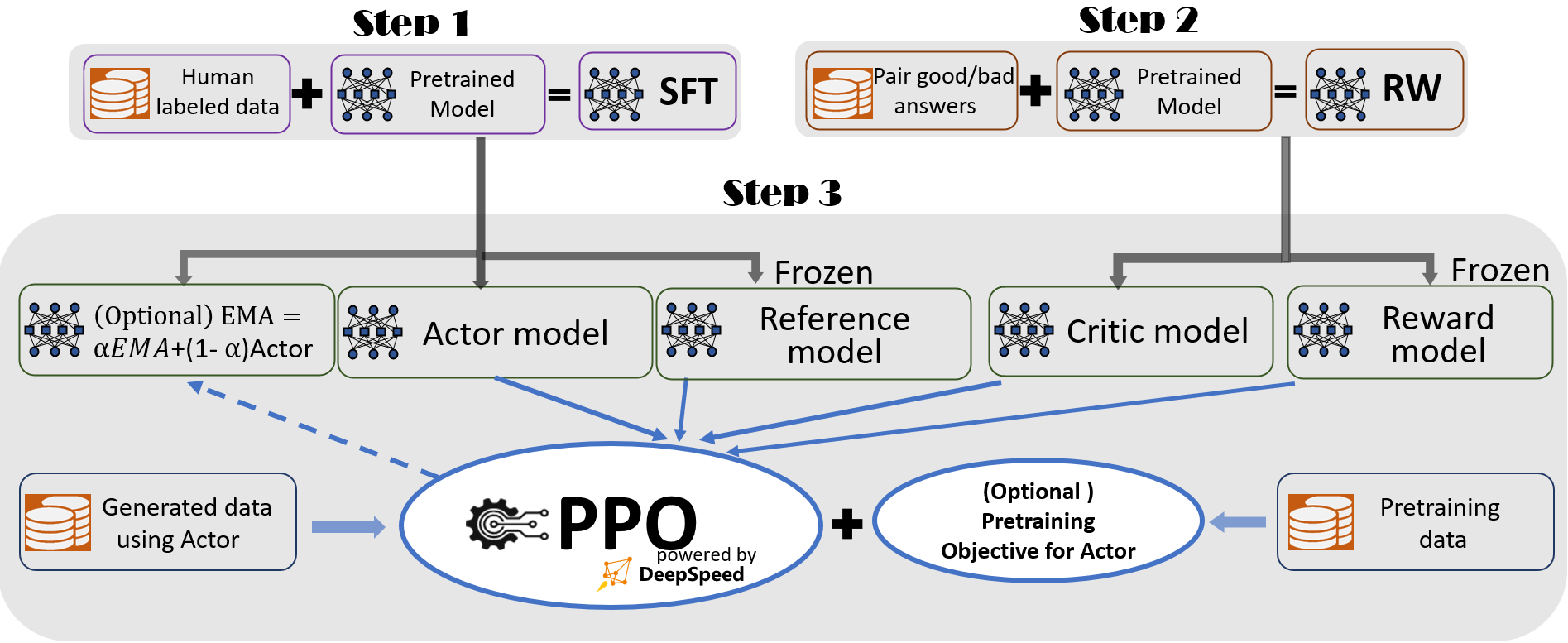}
\caption{The illustration of DeepSpeed-Chat's RLHF training pipeline with optional features.}
\label{fig:figure-1}
\end{figure}

To provide a seamless training experience, we follow InstructGPT and include a full-fledged end-to-end training pipeline in DeepSpeed-Chat as shown in~\fref{fig:figure-1}.

Our pipeline includes three main steps:
\begin{itemize}
\item \textbf{Step 1: Supervised finetuning (SFT)}, where human responses to various queries are carefully selected to finetune the pretrained language models.
\item \textbf{Step 2: Reward model finetuning}, where a separate (usually smaller than the SFT) model (RW) is trained with a dataset that has human-provided rankings of multiple answers to the same query.
\item \textbf{Step 3: RLHF training}, where the SFT model is further finetuned with the reward feedback from the RW model using the Proximal Policy Optimization (PPO)~\cite{schulman2017proximal} algorithm.
\end{itemize}

We provide two additional features in Step 3 to help improve model quality:

\begin{itemize}
\item \textbf{Exponential Moving Average (EMA) collection}, where an EMA based checkpoint can be chosen for the final evaluation.
\item \textbf{Mixture Training}, which mixes the pretraining objective (i.e., the next word prediction) with the PPO objective to prevent regression performance on public benchmarks like SQuAD2.0~\cite{rajpurkar2018know}.
\end{itemize}

The two training features, EMA and Mixed Training, are often omitted by other recent efforts since they can be optional. However, according to InstructGPT, EMA checkpoints generally provide better response quality than conventional final trained model and Mixture Training can help the model retain the pre-training benchmark solving ability. As such, we provide them for users to fully get the training experience as described in InstructGPT and strike for higher model quality.

In addition to being highly consistent with InstructGPT paper~\cite{ouyang2022training}, we also provide convenient features to support researchers and practitioners to train their own RLHF model with multiple data resources:
\begin{itemize}
    \item 
\textbf{Data Abstraction and Blending Capabilities}: DeepSpeed-Chat is able to train the model with multiple datasets for better model quality. It is equipped with (1) an abstract dataset layer to unify the format of different datasets; and (2) data splitting/blending capabilities so that the multiple datasets are properly blended then split across the 3 training stages.
\end{itemize}

To illustrate the effectiveness of our training pipeline, we demonstrate the model quality with multi-round conversation as shown in the experience section.

\section{DeepSpeed Hybrid Engine -- Unified Infrastructure to Power and Optimize RLHF Training}
\label{sec:he}
Step 1 and Step 2 of the instruct-guided RLHF pipeline resemble regular fine-tuning of large models, and they are powered by ZeRO-based optimizations and flexible combination of parallelism strategies in DeepSpeed training to achieve scale and speed.  
Step 3 of the pipeline, on the other hand, is the most complex part to handle in terms of performance implications. 
Each iteration requires efficient processing of two phases a) inference phase for token/experience generation, producing inputs for the training and b) training phase to update the weights of actor and reward models, as well as the interaction and scheduling between them.  
It introduces two major costs: (1) the memory cost, as several copies of the SFT and RW models need to be served throughout stage 3; and (2) the predominant generation phase, which if not accelerated properly, will significantly slow down the entire stage 3. 
Additionally, the two important features we added in Stage 3, including Exponential Moving Average (EMA) collection and Mixture Training, will incur additional memory and training costs.

\begin{figure}
\centering
\includegraphics[width=.85\linewidth]{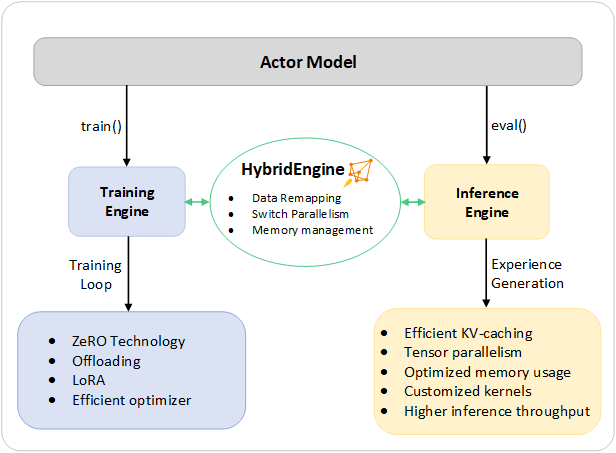}
\caption{ DeepSpeed Hybrid Engine design for accelerating the most time-consuming portion of a RLHF pipeline.}
\label{fig:figure-2}
\end{figure}

To tackle these challenges, we composed the full system capability of DeepSpeed Training and Inference into a unified infrastructure that we call \textbf{Hybrid Engine}. 
It leverages the original DeepSpeed engines for fast training mode while effortlessly applying DeepSpeed inference engine for generation/evaluation mode, providing a significantly faster training system for RLHF training at Stage 3. 
As~\fref{fig:figure-2} shows, the transition between DeepSpeed training and inference engine is seamless: by having the typical eval and train modes enabled for the actor model, when running for inference and training pipeline, DeepSpeed selects its different optimizations to run the model faster and improve the overall system throughput.

During its inference execution for experience generation phase of RLHF training, DeepSpeed Hybrid Engine uses a light-weight memory management system to handle the KV-cache and intermediate results, together with highly optimized inference-adapted kernels and tensor parallelism implementation, to achieve significant boost in throughput (tokens-per-second) compared to the existing solutions.

During the training execution, Hybrid Engine enables memory optimization techniques such as DeepSpeed’s ZeRO family of technologies and Low Rank Adaption (LoRA).   We designed and implemented these system optimizations in a way that they are compatible with each other and can be composed together to deliver the highest training efficiency under the unified Hybrid Engine.

Hybrid Engine can seamlessly change model partitioning across training and inference to support tensor-parallelism based inferencing and ZeRO-based sharding mechanism for training. It can also reconfigure the memory system to maximize memory availability during each of these modes. This allows for improved performance by avoiding memory allocation bottlenecks and supporting large batch sizes. Packed with a spectrum of system technologies from DeepSpeed training and inference, Hybrid Engine pushes the boundary of modern RLHF training and delivers unparalleled scale and system efficiency for RLHF workloads.

\section{DeepSpeed RLHF: Unparalleled Scale and Efficiency via Hybrid Engine
}
\label{sec:scaling}
\subsection{Capability Recap}
As discussed, DeepSpeed-HE is an amalgamation of powerful system technologies for inference and training, architected to achieve excellent scale and efficiency for DeepSpeed-RLHF pipeline across a wide range of hardware, making RLHF training fast, affordable, and easily accessible to AI community.

In terms of efficiency and affordability, as shown in~\tref{tab:table-1}, DeepSpeed-HE can train OPT-13B in just 9 hours and OPT-30B in 18 hours on Azure Cloud for under \$300 and \$600, respectively. 
In terms of speed and scalability, as shown in~\tref{tab:table-2}, even a 13B model can be trained in 1.25 hours and a massive 175B model can be trained in under a day using a 64 GPU cluster. 
And in terms of accessibility and democratization of RLHF, DeepSpeed-HE supports training models with over 13 billion parameters on a single GPU as shown in~\tref{tab:table-3}.

\subsection{Throughput and Model Size Scalability Comparisons with Existing RLHF Systems}
\begin{figure}
\centering
\includegraphics[width=1.0\linewidth]{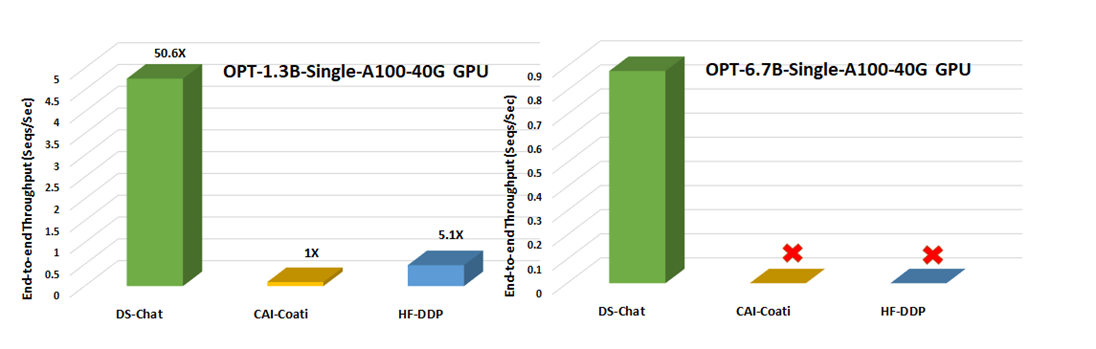}
\caption{Step 3 throughput comparison against two other system frameworks for accelerating RLHF training on a single NVIDIA A100-40G commodity GPU.  No icons represent OOM scenarios.}
\label{fig:figure-3}
\end{figure}

\begin{figure}
\centering
\includegraphics[width=1.0\linewidth]{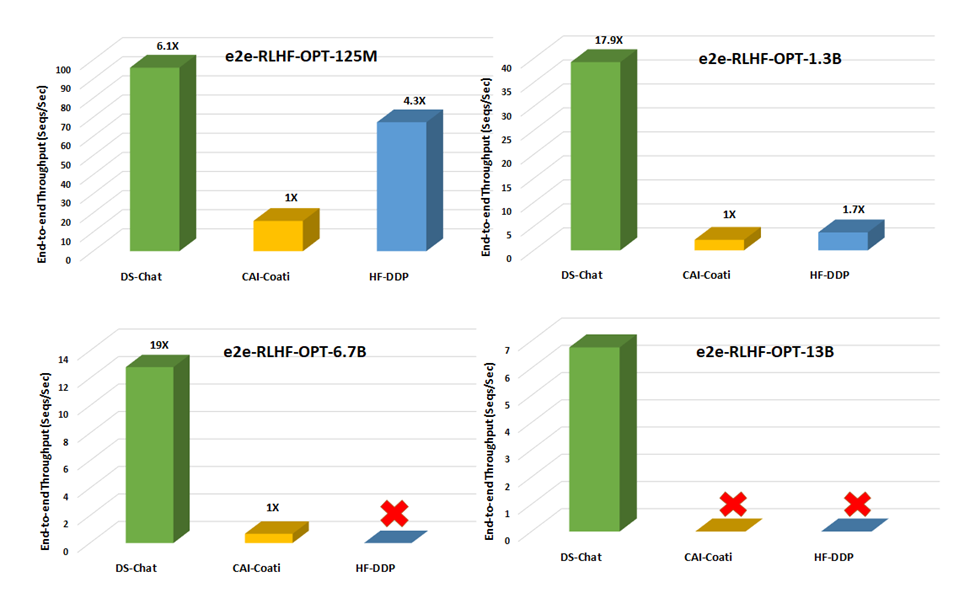}
\caption{End-to-end training throughput comparison for step 3 of the training pipeline (the most time consuming portion) with different model sizes on a single DGX node equipped with 8 NVIDIA A100-40G GPUs. No icons represent OOM scenarios.}
\label{fig:figure-4}
\end{figure}

Compared to other RLHF systems like Colossal-AI~\cite{cai} or HuggingFace~\cite{wolf2019huggingface} powered by native PyTorch~\cite{paszke2019pytorch}, DeepSpeed-RLHF excels in system performance and model scalability:
\begin{itemize}
\item With respect to throughput, DeepSpeed enables over 10x improvement for RLHF training on a single GPU (\fref{fig:figure-3}). On multi-GPU setup, it enables  6 – 19x  speedup over Colossal-AI and 1.4 – 10.5x over HuggingFace DDP (\fref{fig:figure-4}).
\item With respect to model scalability, Colossal-AI can run a max model size of 1.3B on a single GPU and 6.7B on a single A100 40G node, DeepSpeed-HE can run 6.5B and 50B models respectively on the same hardware, up to 7.5x larger.
\end{itemize}
Therefore, with over an order of magnitude higher throughput, DeepSpeed-HE unlocks the ability to train significantly larger actor models under the same latency budget or train models of similar size at over 10x lower cost, compared to existing RLHF systems like Colossal-AI or HuggingFace DDP.

\begin{figure}
\centering
\includegraphics[width=1.0\linewidth]{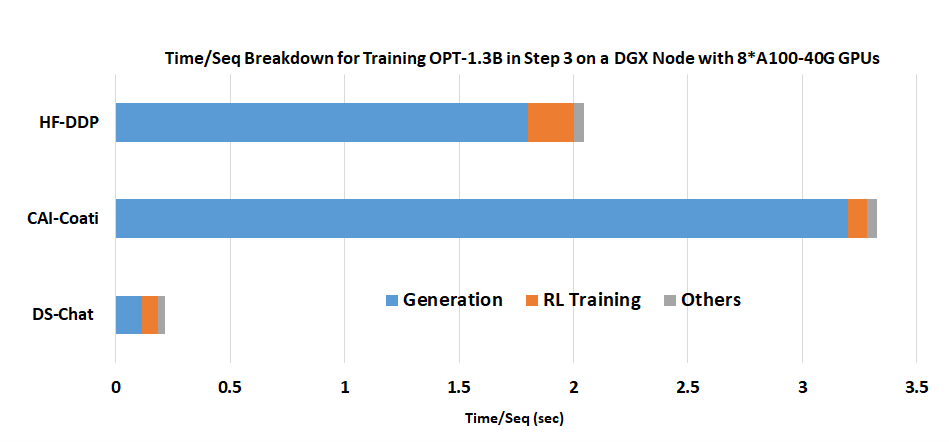}
\caption{Superior generation phase acceleration from DeepSpeed Chat’s Hybrid Engine: A time/sequence breakdown for training OPT-1.3B actor model + OPT-350M reward model on a single DGX node with 8 A100-40G GPUs.}
\label{fig:figure-5}
\end{figure}

This improvement in efficiency stems from DeepSpeed-HE’s ability to accelerate RLHF generation phase of the RLHF processing leveraging DeepSpeed inference optimizations. \fref{fig:figure-5} shows the time breakdown for a 1.3B parameter model at an RLHF training iteration: majority of the time goes to the generation phase. By leveraging high performance inference kernels from DeepSpeed, DeepSpeed-HE can achieve up to 9x throughput improvement during this phase over HuggingFace and 15x over Colossal-AI allowing it to achieve unparallel end-to-end efficiency.

\subsection{Effective Throughput and Scalability Analysis}
\paragraph{(I) Effective Throughput Analysis.}
The effective throughput of DeepSpeed-HE during Stage 3 of the RLHF training depends on the throughput that it achieves during the generation and RL training phases. In our RLHF pipeline, the generation phase comprises approximately 20\% of the total computation while the RL training phase comprises of remaining 80\% (see benchmark settings \url{https://github.com/microsoft/DeepSpeedExamples/tree/master/applications/DeepSpeed-Chat/training/step3_rlhf_finetuning/BenckmarkSetting.md} page for details). 
However, despite having a small proportion, the former can take a large portion of the e2e time as it requires running the actor model once for each of the 256 generated tokens with initial prompt of 256 tokens, making it memory bandwidth bound and difficult to achieve high throughput. In contrast, the RL training phase is compute bound running the reference actor model with just a couple of forward and backward passes with full 512 tokens from both prompt and generation per sample and can achieve good throughput.

\begin{figure}
\centering
\includegraphics[width=.95\linewidth]{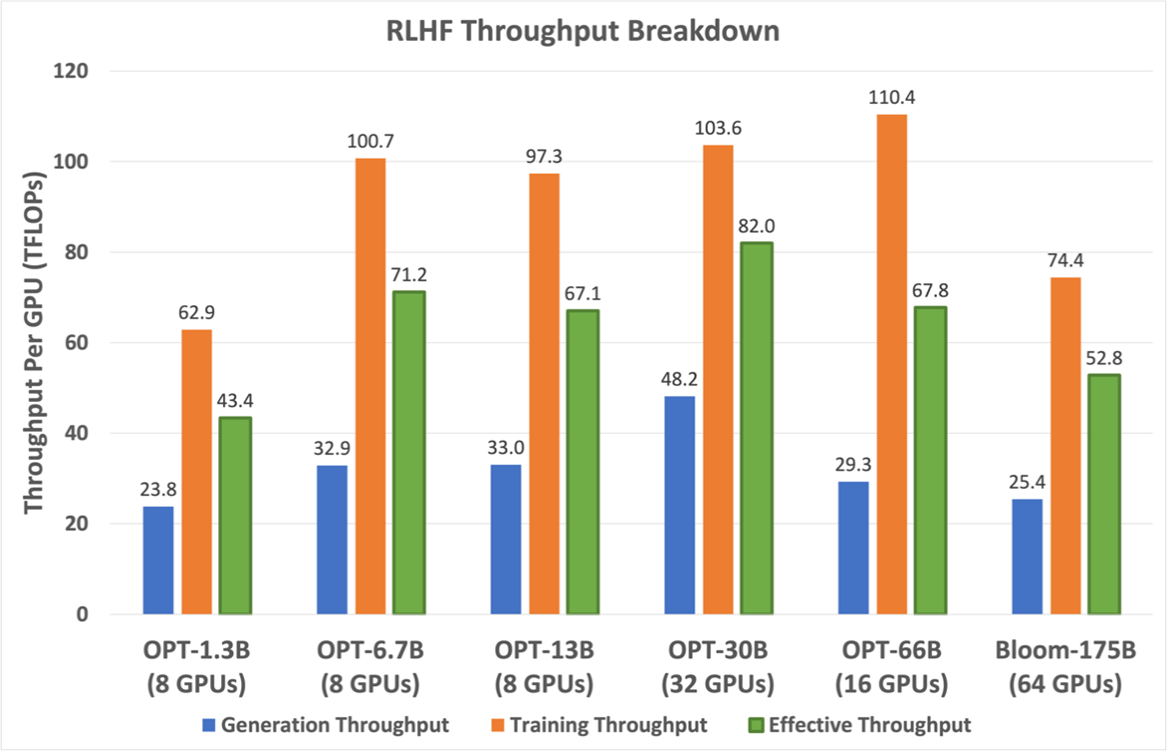}
\caption{RLHF Generation, training, and effective throughput with DeepSpeed-HE for different model sizes, at the GPU count that maximizes efficiency.}
\label{fig:figure-6}
\end{figure}

To maximize the effective throughput, DeepSpeed-HE optimizes both phases. First, it uses the largest batch size possible to get higher efficiency on both phases. Second, during the generation phase, it leverages high-performance transformer kernels to maximize GPU memory bandwidth utilization when the model fits in single GPU memory, and leverage tensor-parallelism (TP) when it does not. Using TP in the generation phase instead of ZeRO to fit the model reduces the inter-GPU communication and maintains high GPU memory bandwidth utilization.

\fref{fig:figure-6} shows the best achievable effective throughput for DeepSpeed-HE in terms of TFlops/GPU for model sizes ranging from 1.3B to 175B. It also shows the throughput achieved by each of the generation and training phases. 
DeepSpeed-HE is the most efficient for models in the range 6.7B-66B. 
Going beyond this range to 175B, the throughput drops due to the limited memory to support larger batch sizes, while still achieving 1.2x better efficiency than the small 1.3B model. 
The per-GPU throughput of these gigantic models could improve further when we scale them to more GPUs with more memory available for larger batch sizes.

Furthermore, we would like to point out that our effective performance is 19x higher than existing systems, as shown in~\fref{fig:figure-4}, which suggests that they are operating at lower than 5\% of the peak. This demonstrates the challenge of optimizing RLHF workloads as well as the effectiveness of our system despite the challenge.

\begin{figure}
\centering
\includegraphics[width=1.0\linewidth]{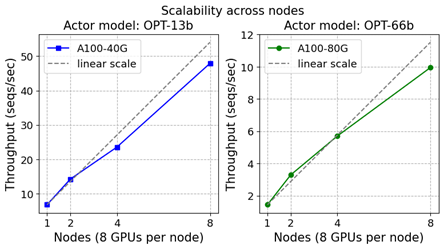}
\caption{Scalability for training 13B (left) and 66B (right) actor model+350M reward model on an increasing number of DGX nodes with 8 A100-40/80G GPUs}
\label{fig:figure-7}
\end{figure}

\paragraph{(II) Scalability Analysis} The best effective throughput for different model sizes is achieved at different GPU count. 
This is in part because some of the larger model sizes require more memory to run. 
However, a large part of this behavior stems from DeepSpeed-HE’s scalability properties that we discuss next.

\fref{fig:figure-7} shows that DeepSeed-RLHF has achieved good scaling overall on up to 64 GPUs. 
However, if we look more closely, it shows that DeepSpeed-RLHF training achieves super-linear scaling at small scale, followed by near linear or sub-linear scaling at larger scales. 
This is due to interaction between memory availability and max global batch size.

As DeepSpeed-HE is powered by ZeRO-based technology~\cite{rajbhandari2020zero} for training, it allows model states to be partitioned across the available GPUs. 
As a result, the memory consumption per GPU reduces with the increase in the number of GPUs, allowing DeepSpeed-HE to support a larger batch per GPU resulting in super-linear scaling. 
However, at large scale, while the available memory continues to increase, the maximum global batch size (1024, in our case, with a sequence length of 512) limits the batch size per GPU, resulting in near-linear or sub-linear scaling.
As a result, for a given max global batch size, DeepSpeed-HE achieves the best throughput and cost efficiency at the boundary of super-linear and sub-linear scalability, and the exact point is mostly determined by the largest batch size that can be run per GPU as the function of available memory and global batch size.

\section{Release: Try DeepSpeed Chat Today!}
We are very excited to share that DeepSpeed-Chat is now open-sourced and available to the AI community.

\begin{itemize}
\item To get started, please visit our github page for DeepSpeed-Chat: GitHub Landing Page (\url{https://github.com/microsoft/DeepSpeedExamples/tree/master/applications/DeepSpeed-Chat})

\item We will continue to improve DeepSpeed-Chat with your feedback and support. Our roadmap (\url{https://github.com/microsoft/DeepSpeedExamples/tree/master/applications/DeepSpeed-Chat/README.md#-deepspeed-chats-roadmap-}) shows currently supported features as well as ones that are planned for future.
\end{itemize}

DeepSpeed-Chat is part of the bigger DeepSpeed ecosystem comprising of a multitude of Deep Learning systems and modeling technologies. To learn more,
\begin{itemize}
\item  Please visit our website (\url{https://www.deepspeed.ai/}) for detailed blog posts, tutorials, and helpful documentation.
\item You can also follow us on our English Twitter (\url{https://twitter.com/MSFTDeepSpeed}), Japanese Twitter (\url{https://twitter.com/MSFTDeepSpeedJP}), and Chinese Zhihu (\url{https://www.zhihu.com/people/deepspeed}) for latest news on DeepSpeed.
\end{itemize}
DeepSpeed welcomes your contributions! We encourage you to report issues, contribute PRs, and join discussions on the DeepSpeed GitHub (\url{https://github.com/microsoft/DeepSpeed/}) page. Please see our contributing guide (\url{https://github.com/microsoft/DeepSpeed/blob/master/CONTRIBUTING.md}) for more details. We are open to collaborations with universities, research labs, companies, such as those working together on deep learning research, applying DeepSpeed to empower real-world AI models and applications, and so on.

\section*{Contributions}
\label{sec:contribution}

\noindent \textbf{ZY}: Full engagement. 
\noindent \textbf{RYA}: Hybrid engine.
\noindent \textbf{OR}: DeepSpeed ZeRO feature adaptation.
\noindent \textbf{SR}: System support and blog contribution.
\noindent \textbf{XW}: Training pipeline and bench-marking support.
\noindent \textbf{AAA}: Software support and post-release debugging.
\noindent \textbf{JR}: Software support and post-release debugging.
\noindent \textbf{MZ}: Training pipeline and system support, post-release debugging.
\noindent \textbf{CL}: Data support and post-release debugging.
\noindent \textbf{CH}: System support.
\noindent \textbf{ZZ}: Benchmarking.
\noindent \textbf{MW}: Software support and post-release debugging.
\noindent \textbf{MS}: Post-release debugging.
\noindent \textbf{LK}: Post-release debugging. 
\noindent \textbf{HQ}: System support.
\noindent \textbf{MT}: System support.
\noindent \textbf{SC}: Software support.
\noindent \textbf{SLS}: System support, blog and tutorial contribution.
\noindent \textbf{YH}: Team lead.

\section*{Acknowledgment}
We thank the entire DeepSpeed team for their contributions on developing, debugging, testing, and releasing the DeepSpeed-Chat software.

\bibliographystyle{unsrt}
\bibliography{references}

\begin{thebibliography}{10}

\bibitem{chatgpt}
OpenAI.
\newblock Chatgpt.
\newblock \url{https://openai.com/blog/chatgpt}, 2022.

\bibitem{chatllama}
ChatLLaMa Authors.
\newblock Chatllama.
\newblock \url{https://github.com/juncongmoo/chatllama}, 2023.

\bibitem{alpaca}
Rohan Taori, Ishaan Gulrajani, Tianyi Zhang, Yann Dubois, Xuechen Li, Carlos
  Guestrin, Percy Liang, and Tatsunori~B. Hashimoto.
\newblock Stanford alpaca: An instruction-following llama model.
\newblock \url{https://github.com/tatsu-lab/stanford_alpaca}, 2023.

\bibitem{zheng2023judging}
Lianmin Zheng, Wei-Lin Chiang, Ying Sheng, Siyuan Zhuang, Zhanghao Wu, Yonghao
  Zhuang, Zi~Lin, Zhuohan Li, Dacheng Li, Eric.~P Xing, Hao Zhang, Joseph~E.
  Gonzalez, and Ion Stoica.
\newblock Judging llm-as-a-judge with mt-bench and chatbot arena, 2023.

\bibitem{dolly}
Databricks.
\newblock Databricks-dolly.
\newblock
  \url{https://www.databricks.com/blog/2023/03/24/hello-dolly-democratizing-magic-chatgpt-open-models.html?scid=7018Y000001Fi1CQAS&utm_medium=paid+search&utm_source=google&utm_campaign=17107065832&utm_adgroup=150868748114&utm_content=blog&utm_offer=hello-dolly-democratizing-magic-chatgpt-open-models.html&utm_ad=661606835947&utm_term=databricks%20dolly&gclid=Cj0KCQjwiIOmBhDjARIsAP6YhSV89V2agFl3zFuWiZiV1N3IVNhZWr8pGtXVxXrlkuPHlW3cXbGfiHsaAmIDEALw_wcB},
  2023.

\bibitem{wolf2019huggingface}
Thomas Wolf, Lysandre Debut, Victor Sanh, Julien Chaumond, Clement Delangue,
  Anthony Moi, Pierric Cistac, Tim Rault, R{\'e}mi Louf, Morgan Funtowicz,
  et~al.
\newblock Huggingface's transformers: State-of-the-art natural language
  processing.
\newblock {\em arXiv preprint arXiv:1910.03771}, 2019.

\bibitem{ouyang2022training}
Long Ouyang, Jeffrey Wu, Xu~Jiang, Diogo Almeida, Carroll Wainwright, Pamela
  Mishkin, Chong Zhang, Sandhini Agarwal, Katarina Slama, Alex Ray, et~al.
\newblock Training language models to follow instructions with human feedback.
\newblock {\em Advances in Neural Information Processing Systems},
  35:27730--27744, 2022.

\bibitem{stiennon2020learning}
Nisan Stiennon, Long Ouyang, Jeffrey Wu, Daniel Ziegler, Ryan Lowe, Chelsea
  Voss, Alec Radford, Dario Amodei, and Paul~F Christiano.
\newblock Learning to summarize with human feedback.
\newblock {\em Advances in Neural Information Processing Systems},
  33:3008--3021, 2020.

\bibitem{hu2021lora}
Edward~J Hu, Yelong Shen, Phillip Wallis, Zeyuan Allen-Zhu, Yuanzhi Li, Shean
  Wang, Lu~Wang, and Weizhu Chen.
\newblock Lora: Low-rank adaptation of large language models.
\newblock {\em arXiv preprint arXiv:2106.09685}, 2021.

\bibitem{zhang2022opt}
Susan Zhang, Stephen Roller, Naman Goyal, Mikel Artetxe, Moya Chen, Shuohui
  Chen, Christopher Dewan, Mona Diab, Xian Li, Xi~Victoria Lin, et~al.
\newblock Opt: Open pre-trained transformer language models.
\newblock {\em arXiv preprint arXiv:2205.01068}, 2022.

\bibitem{schulman2017proximal}
John Schulman, Filip Wolski, Prafulla Dhariwal, Alec Radford, and Oleg Klimov.
\newblock Proximal policy optimization algorithms.
\newblock {\em arXiv preprint arXiv:1707.06347}, 2017.

\bibitem{rajpurkar2018know}
Pranav Rajpurkar, Robin Jia, and Percy Liang.
\newblock Know what you don't know: Unanswerable questions for squad.
\newblock {\em arXiv preprint arXiv:1806.03822}, 2018.

\bibitem{cai}
Colossal~AI Authors.
\newblock Colossal ai.
\newblock \url{https://github.com/hpcaitech/ColossalAI }, 2022.

\bibitem{paszke2019pytorch}
Adam Paszke, Sam Gross, Francisco Massa, Adam Lerer, James Bradbury, Gregory
  Chanan, Trevor Killeen, Zeming Lin, Natalia Gimelshein, Luca Antiga, et~al.
\newblock Pytorch: An imperative style, high-performance deep learning library.
\newblock {\em Advances in neural information processing systems}, 32, 2019.

\bibitem{rajbhandari2020zero}
Samyam Rajbhandari, Jeff Rasley, Olatunji Ruwase, and Yuxiong He.
\newblock Zero: Memory optimizations toward training trillion parameter models.
\newblock In {\em SC20: International Conference for High Performance
  Computing, Networking, Storage and Analysis}, pages 1--16. IEEE, 2020.

\end{thebibliography}

\newpage
\appendix

\end{document}